\documentclass[10pt,twocolumn,letterpaper]{article}

\usepackage{iccv}
\usepackage{times}
\usepackage{epsfig}
\usepackage{graphicx}
\usepackage{amsmath}
\usepackage{amssymb}
\usepackage{placeins}
\usepackage{booktabs}

\newcommand{\fig}[1]{Figure~\ref{fig:#1}}

\newcommand{\eq}[1]{(\ref{eq:#1})}

\usepackage[pagebackref=true,breaklinks=true,letterpaper=true,colorlinks,bookmarks=false]{hyperref}

\iccvfinalcopy %

\def\iccvPaperID{5696} %

\ificcvfinal\fi
\begin{document}

\title{Learnable Triangulation of Human Pose}

\author{Karim Iskakov $^1$ \quad Egor Burkov $^{1,2}$ \quad Victor Lempitsky $^{1,2}$ \quad  Yury Malkov $^1$ 
\vspace{1.5mm}\\
$^1 $Samsung AI Center, Moscow \quad $^2 $Skolkovo Institute of Science and Technology, Moscow\\
{\tt\small \{k.iskakov, e.burkov, v.lempitsky, y.malkov\}@samsung.com}
}

\maketitle
\begin{abstract}
We present two novel solutions for multi-view 3D human pose estimation based on new learnable triangulation methods that combine 3D information from multiple 2D views. The first (baseline) solution is a basic differentiable algebraic triangulation with an addition of confidence weights estimated from the input images. The second solution is based on a novel method of volumetric aggregation from intermediate 2D backbone feature maps. The aggregated volume is then refined via 3D convolutions that produce final 3D joint heatmaps and allow modelling a human pose prior. Crucially, both approaches are end-to-end differentiable, which allows us to directly optimize the target metric. We demonstrate transferability of the solutions across datasets and considerably improve the multi-view state of the art on the Human3.6M dataset. Video demonstration, annotations and additional materials will be posted on our project page\footnote{\url{https://saic-violet.github.io/learnable-triangulation}}.
\end{abstract}

\section{Introduction}
3D human pose estimation is one of the fundamental problems in computer vision, with applications in sports, action recognition, computer-assisted living, human-computer interfaces, special effects, and telepresence. To  date, most of the efforts in the community are focused on \textit{monocular} 3D pose estimation. Despite a lot of recent progress, the problem of in-the-wild monocular 3D human pose estimation is far from being solved. Here, we consider a simpler yet still challenging problem of multi-view 3D human pose estimation.

There are at least two reasons why multi-view human pose estimation is interesting. First, multi-view pose estimation is arguably the best way to obtain ground truth for monocular 3D pose estimation~\cite{Joo_2017_TPAMI, yu2018humbi} in-the-wild. This is because the competing techniques such as marker-based motion capture~\cite{moeslund2006survey} and visual-inertial methods~\cite{von2016human} have certain limitations such as inability to capture rich pose representations (e.g.\ to estimate hands pose and face pose alongside limb pose) as well as various clothing limitations. The downside is, previous works that used multi-view triangulation for constructing datasets relied on excessive, almost impractical number of views to get the 3D ground truth of sufficient quality~\cite{Joo_2017_TPAMI, yu2018humbi}. This makes the collection of new in-the-wild datasets for 3D pose estimation very challenging and calls for the reduction of the number of views needed for accurate triangulation.

The second motivation to study multi-view human pose estimation is that, in some cases, it can be used directly to track human pose in real-time for the practical end purpose. This is because multi-camera setups are becoming progressively available in the context of various applications, such as sports or computer-assisted living. Such practical multi-view setups rarely go beyond having just a few views. At the same time, in such a regime, modern multi-view methods have accuracy comparable to well-developed monocular methods~\cite{Tome2018, Pavlakos2017, Kadkhodamohammadi2018,sun2018integral, pavllo:videopose3d:2018}. Thus, improving the accuracy of multi-view pose estimation from few views is an important challenge with direct practical applications.

In this work, we argue that given its importance, the task of multi-view pose estimation has received disproportionately little attention. We propose and investigate two simple and related methods for multi-view human pose estimation. Behind both of them lies the idea of \textit{learnable} triangulation, which allows us to dramatically reduce the number of views needed for accurate estimation of 3D pose. During learning, we either use marker based motion capture ground truth or ``meta''-ground truth obtained from the excessive number of views. The methods themselves are as follows: (1) a simpler approach based on algebraic triangulation with learnable camera-joint confidence weights, and (2) a more complex volumetric triangulation approach based on dense geometric aggregation of information from different views that allows modelling a human pose prior. Crucially, both of the proposed solutions are fully differentiable, which permits end-to-end training.

Below, we review related work in monocular and multi-view human pose estimation, and then discuss the details of the new learnable triangulation methods. In the experimental section, we perform an evaluation on the popular Human3.6M~\cite{h36m_pami} and CMU Panoptic~\cite{Joo_2017_TPAMI} datasets, demonstrating state-of-the-art accuracy of the proposed methods and and their ability of cross-dataset generalization.

\section{Related work}
\label{sect:related}

\paragraph{Single view 3D pose estimation.}
Current state-of-the-art solutions for the monocular 3D pose estimation can be divided into two sub-categories. The first category is using high quality 2D pose estimation engines with subsequent separate lifting of the 2D coordinates to 3D via deep neural networks (either fully-connected, convolutional or recurrent). This idea was popularized in \cite{martinez2017simple} and offers several advantages: it is simple, fast, can be trained on motion capture data (with skeleton/view augmentations) and allows switching 2D backbones after training. Despite known ambiguities inherent to this family of methods (i.e. orientation of arms' joints in current skeleton models), this paradigm is adopted in the current multi-frame state of the art~\cite{pavllo:videopose3d:2018} on the Human3.6M benchmark~\cite{h36m_pami}.

The second option is to infer the 3D coordinates directly from the images using convolutional neural networks. The present best solutions use volumetric representations of the pose, with current single-frame state-of-the-art results on Human3.6M~\cite{h36m_pami}, namely \cite{sun2018integral}. %

\paragraph{Multi-view view 3D pose estimation.} Studies of multi-view 3D human pose estimation are generally aimed at getting the ground-truth annotations for the monocular 3D human pose estimation \cite{Rhodin2018, Joo_2017_TPAMI}. The work~\cite{Kadkhodamohammadi2018} proposed concatenating joints' 2D coordinates from all views into a single batch as an input to a fully connected network that is trained to predict the global 3D joint coordinates. This approach can efficiently use the information from different views and can be trained on motion capture data. However, the method is by design unable to transfer the trained models to new camera setups, while the authors show that the approach is prone to strong over-fitting.

Few works used volumetric pose representation in multi-view setups\cite{Pavlakos2017,Joo_2017_TPAMI}. Specifically, \cite{Joo_2017_TPAMI} utilized unprojection of 2D keypoint probability heatmaps (obtained from a pretrained 2D keypoint detector) to volume with subsequent non-learnable aggregation. Our work differs in two ways. First, we process information inside the volume in a learnable way. Second, we train the network end-to-end, thus adjusting the 2D backbone and alleviating the need for interpretability of the 2D heatmaps. This allows to transfer several self-consistent pose hypotheses from 2D detectors to the volumetric aggregation stage (which was not possible with previous designs).

The work~\cite{Tome2018} used a multi-stage approach with an external 3D pose prior~\cite{tome2017lifting} to infer the 3D pose from 2D joints' coordinates. During the first stage, images from all views were passed through the backbone convolutional neural network to obtain 2D joints' heatmaps. The positions of maxima in the heatmaps were jointly used to infer the 3D pose via optimizing latent coordinates in 3D pose prior space. In each of the subsequent stages, 3D pose was reprojected back to all camera views and fused with predictions from the previous layer (via a convolutional network). Next, the 3D pose was re-estimated from the positions of heatmap maxima, and the process repeated. Such procedure allowed correcting the predictions of 2D joint heatmaps via indirect holistic reasoning on a human pose.
In contrast to our approach, in \cite{Tome2018} there is no gradient flow from the 3D predictions to 2D heatmaps and thus no direct signal to correct the prediction of 3D coordinates. %

\newcommand{\vect}[1]{\mathbf{#1}}
\newcommand{\bm}[1]{{#1}}

\section{Method}

\begin{figure*}[t]
    \centering
    \includegraphics[width=0.85\textwidth]{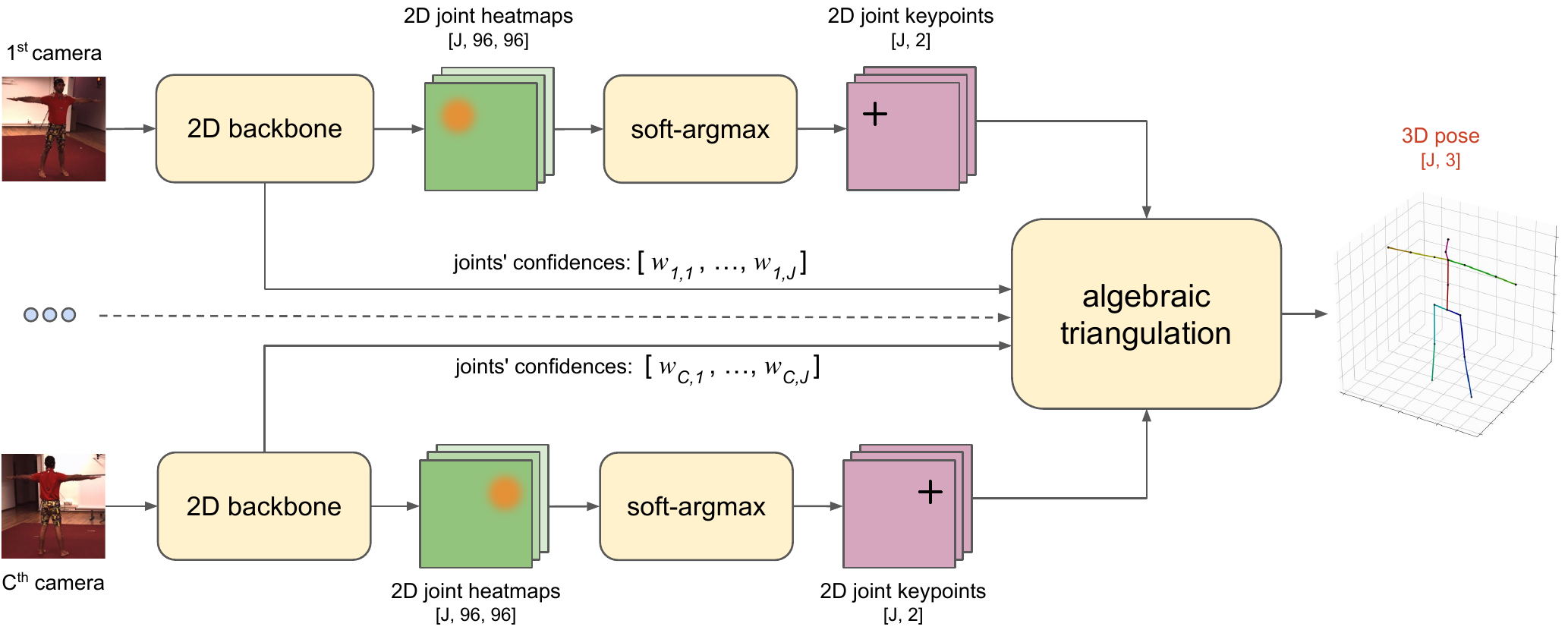}
    \caption{Outline of the approach based on algebraic triangulation with learned confidences. The input for the method is a set of RGB images with known camera parameters. The 2D backbone produces the joints' heatmaps and camera-joint confidences. The 2D positions of the joints are inferred from 2D joint heatmaps by applying soft-argmax. The 2D positions together with the confidences are passed to the algebraic triangulation module that outputs the triangulated 3D pose. All blocks allow backpropagation of the gradients, so the model can be trained end-to-end.}
    \label{fig:triang}
\end{figure*}

Our approach assumes we have synchronized video streams from $C$ cameras with known projection matrices $P_c$ capturing performance of a single person in the scene. We aim at estimating the global 3D positions $\vect{y}_{j,t}$ of a fixed set of human joints with indices $j\in(1..J)$ at timestamp $t$. For each timestamp the frames are processed independently  (i.e. without using temporal information), thus we omit the index $t$ for clarity.

For each frame, we crop the images using the bounding boxes either estimated by available off-the-shelf 2D human detectors or from ground truth (if provided). Then we feed the cropped images $I_{c}$ into a deep convolutional neural network backbone based on the "simple baselines" architecture~\cite{xiao2018simple}. 

The convolutional neural network backbone with learnable weights $\theta$ consists of a ResNet-152 network (output denoted by $g_{\theta}$), followed by a series of transposed convolutions that produce intermediate heatmaps (the output denoted by $f_{\theta}$) and a $1\times1$ - kernel convolutional neural network that transforms the intermediate heatmaps to interpretable joint heatmaps (output denoted by $h_\theta$; the number of output channels is equal to the number of joints $J$).
In the two following sections we describe two different methods to infer joints' 3D coordinates by aggregating information from multiple views.

\paragraph{Algebraic triangulation approach.}
In the algebraic triangulation baseline we process each joint $j$ independently of each other. The approach is built upon triangulating the 2D positions obtained from the $j$-joint's backbone heatmaps from different views: $H_{c,j}=h_\theta(I_{c})_j$ (\fig{triang}). To estimate the 2D positions we first compute the softmax across the spatial axes: 
\begin{equation}
      H'_{c,j}={\text{exp}(\alpha H_{c,j})} / 
      \Big({\sum\limits_{{{r}_{x}}=1}^{W}{\sum\limits_{{{r}_{y}}=1}^{H} \text{exp}(\alpha H_{c,j}(\vect{r}))}} \label{eq:softmax1}\Big),
\end{equation}
where parameter $\alpha$ is discussed below. Then we calculate the 2D positions of the joints as the center of mass of the corresponding heatmaps (so-called soft-argmax operation):
\begin{equation}
    \vect{x}_{c,j}= \sum\limits_{{{r}_{x}}=1}^{W}{\sum\limits_{{{r}_{y}}=1}^{H}{\vect{r}\cdot( {H'}_{c,j}(\vect{r})}}) 
\end{equation}

An important feature of soft-argmax is that rather than getting the index of the maximum, it allows the gradients to flow back to heatmaps $H_{c}$ from the output 2D position of the joints $\vect{x}$.  Since the backbone was pretrained using a loss other than soft-argmax (MSE over heatmaps without softmax~\cite{sun2018integral}), we adjust the heatmaps via multiplying them by an 'inverse temperature' parameter $\alpha=100$ in \eq{softmax1}, so at the start of the training the soft-argmax gives an output close to the positions of the maximum.

To infer the 3D positions of the joints from their 2D estimates $\vect{x}_{c,j}$ we use a linear algebraic triangulation approach~\cite{hartley2003multiple}. The method reduces the finding of the 3D coordinates of a joint $\vect{y}_j$ to solving the overdetermined system of equations on homogeneous 3D coordinate vector of the joint $\vect{\tilde{y}}$: 
\begin{equation}
\bm{A_j}\tilde{\vect{y}}_j = 0 \label{eq:algtriag1},
\end{equation}
where $\bm{A_j}\in \mathbb{R}^{(2C, 4)}$ is a matrix composed of the components from the full projection matrices and $\vect{x}_{c,j}$ (see \cite{hartley2003multiple} for full details). 

A na\"ive triangulation algorithm assumes that the joint coordinates from each view are independent of each other and thus all make comparable contributions to the triangulation. However, on some views the 2D position of the joints cannot be estimated reliably (e.g. due to joint occlusions), leading to unnecessary degradation of the final triangulation result. This greatly exacerbates the tendency of methods that optimize algebraic reprojection error to pay uneven attention to different views. The problem can be dealt with by applying RANSAC together with the Huber loss (used to score reprojection errors corresponding to inliers). However, this has its own drawbacks. E.g.\ using RANSAC may completely cut off the gradient flow to the excluded cameras.

\begin{figure*}
    \centering
    \includegraphics[width=0.9\textwidth]{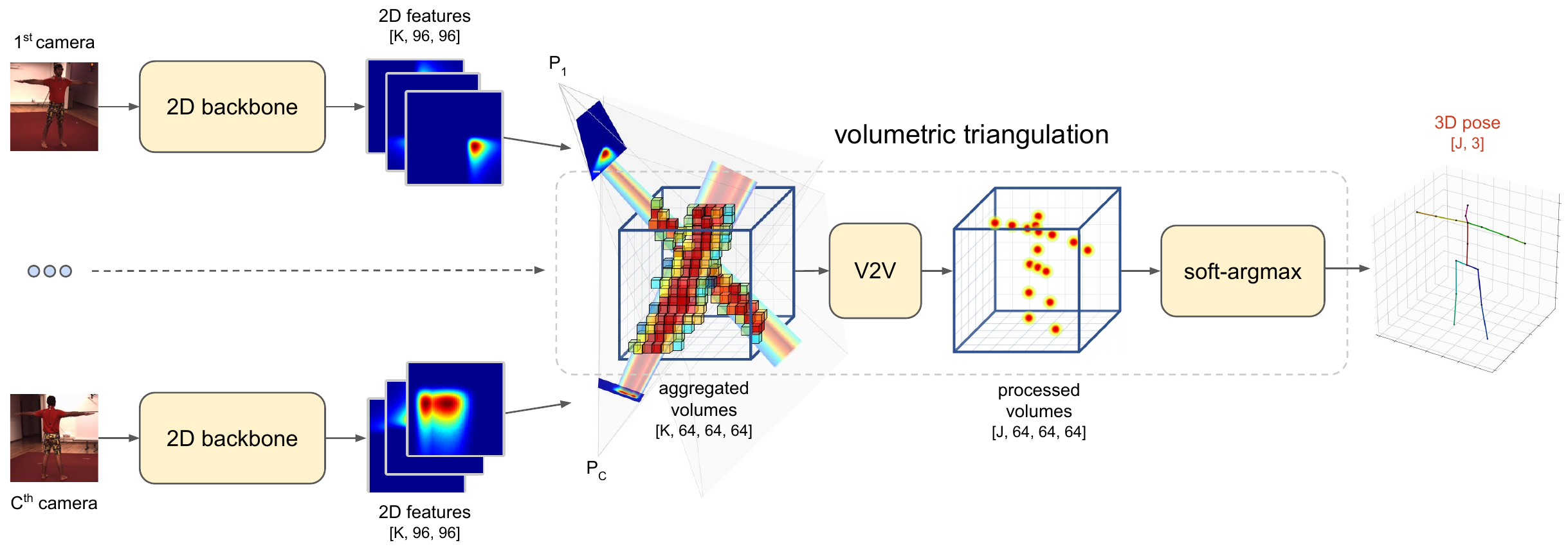}
    \caption{Outline of the approach based on volumetric triangulation. The input for the method is a set of RGB images with known camera parameters. The 2D backbone produces intermediate feature maps that are unprojected into volumes with subsequent aggreagation to a fixed size volume. The volume is passed to a 3D convolutional neural network that outputs the interpretable 3D heatmaps. The output 3D positions of the joints are inferred from 3D joint heatmaps by computing soft-argmax. All blocks allow backpropagation of the gradients, so the model can be trained end-to-end.}
    \label{fig:vol}
\end{figure*}

To address the aforementioned problems, we add \textit{learnable} weights $w_c$ to the coefficients of the matrix corresponding to different views:
\begin{equation}
(\vect{w}_j\circ\bm{A_j})\tilde{\vect{y}}_j=0, \label{eq:algtriag2}
\end{equation}
where $\vect{w}_j=(w_{1,j},w_{1,j},w_{2,j},w_{2,j},...,,w_{C,j},w_{C,j})$; $\circ$ denotes the Hadamard product (i.e. $i$-th row of matrix $\bm{A}$ is multiplied by $i$-th element of vector $\vect{w}$). The weights $w_{c,j}$ are estimated by a convolutional network $q^\phi$ with learnable parameters $\phi$ (comprised of two convolutional layers, global average pooling and three fully-connected layers), applied to the intermediate output of the backbone:
\begin{equation}
w_{c,j}=\big(q^\phi (g^{\theta}(I_c))  \label{eq:algtriagcoeff}\big)_j
\end{equation}
This allows the contribution of the each camera view to be controlled by the neural network branch that is learned jointly with the backbone joint detector.

The equation \eq{algtriag2} is solved via differentiable Singular Value Decomposition of the matrix $\bm{B}=\bm{\bm{U}\bm{D}\bm{V^T}}$, from which $\tilde{\vect{y}}$ is set as the last column of $\bm{V}$.
The final non-homogeneous value of $\vect{y}$ is obtained by dividing the homogeneous 3D coordinate vector $\tilde{\vect{y}}$ by its fourth coordinate: $\vect{y}=\tilde{\vect{y}} / (\tilde{\vect{y}})_4$.

\paragraph{Volumetric triangulation approach.}
The main drawback of the baseline algebraic triangulation approach is that the images $I_c$ from different cameras are processed independently from each other, so there is no easy way to add a 3D human pose prior and no way to filter out the cameras with wrong projection matrices. 

To solve this problem we propose to use a more complex and powerful triangulation procedure. We unproject the feature maps produced by the 2D backbone into 3D volumes (see \fig{vol}). This is done by filling a 3D cube around the person via projecting output of the 2D network along projection rays inside the 3D cube. The cubes obtained from multiple views are then aggregated together and processed. For such volumetric triangulation approach, the 2D output does not have to be interpretable as joint heatmaps, thus, instead of unprojecting $H_c$ themselves, we use the output of a trainable single layer convolutional neural network $o^\gamma$ with $1\times1$ kernel and $K$ output channels (the weights of this layer are denoted by $\gamma$) applied to the input from the backbone intermediate heatmaps $f^{\theta}(I_c)$:
\begin{equation}
M_{c,k}=o^\gamma(f^{\theta}(I_c))_k  \label{eq:vol_input}
\end{equation}
To create the volumetric grid, we place a $L \times L \times L$ - sized 3D bound box in the global space around the human pelvis (the position of the pelvis is estimated by the algebraic triangulation baseline described above, $L$ denotes the size of the box in meters) with the Y-axis perpendicular to the ground and a random orientation of the X-axis. We discretize the bounding box by a volumetric cube $V^{\text{coords}} \in \mathbb{R}^{64,64,64,3}$, filling it with the global coordinates of the center of each voxel (in a similar way to \cite{Joo_2017_TPAMI}).

For each view, we then project the 3D coordinates in $V^{\text{coords}}$ to the plane: $V_c^{\text{proj}}=P_c V^{\text{coords}}$ (note that $V_c^{\text{proj}} \in \mathbb{R}^{64,64,64,2}$) and fill a cube $V_c^{\text{view}}\in \mathbb{R}^{64,64,64,K}$ by bilinear sampling~\cite{jaderberg2015spatial} from the maps $M_{c,k}$ of the corresponding camera view using 2D coordinates in $V_c^{\text{proj}}$: 
\begin{equation}
V_{c,k}^{\text{view}}= M_{c,k}\{V_c^{\text{proj}}\},
\end{equation}
where $\{\cdot\}$ denotes bilinear sampling.
We then aggregate the volumetric maps from all views to form an input to the further processing that does not depend on the number of camera views. We study three diverse methods for the aggregation:
\begin{enumerate}
\item Raw summation of the voxel data: 
\begin{equation}
V^{\text{input}}_k=\sum\limits_c{V_{c,k}^{\text{view}}}   \label{eq:vol_aggr_sum}
\end{equation}
\item Summation of the voxel data with normalized confidence multipliers $d_c$ (obtained similarly to $w_c$ using a branch attached to backbone): 
\begin{equation}
V^{\text{input}}_k=\sum\limits_c{\Big(d_c \cdot V_{c,k}^{\text{view}}\Big) / \sum\limits_c{d_c}} \label{eq:vol_aggr_conf}
\end{equation}
\item Calculating a relaxed version of maximum. Here, we first compute the softmax for each individual voxel $V_c^{\text{view}}$ across all cameras, producing the volumetric coefficient distribution $V^w_{c,k}$ with the role similar to scalars $d_c$:
\begin{equation}
  V^w_{c,k}=\text{exp}(V_{c,k}^{\text{view}})/\sum\limits_c{\text{exp}(V_{c,k}^{\text{view}})} \label{eq:vol_aggr_soft_softmax}
\end{equation}
Then, the voxel maps from each view are summed with the volumetric coefficients $V^w_c$:
\begin{equation}
V^{\text{input}}_k=\sum\limits_c{V^w_{c,k} \circ V_c^{\text{view}} } \label{eq:vol_aggr_soft}
\end{equation}
\end{enumerate}

Aggregated volumetric maps are then fed into a learnable volumetric convolutional neural network $u^{\nu}$ (with weights denoted by $\nu$), with architecture similar to V2V~\cite{Moon_2018_CVPR_V2V-PoseNet}, producing the interpretable 3D-heatmaps of the output joints:
\begin{equation}
{V}_{j}^{\text{output}}=(u^{\nu}( V^{\text{input}}))_j
\end{equation}

Next, we compute softmax of ${V}_{j}^{\text{output}}$ across the spatial axes (similar to \eq{softmax1}): 
\begin{equation}
{V'}_{j}^{\text{output}}={\text{exp}( {V}_{j}^{\text{output}} )} / 
      \Big({\sum\limits_{{{r}_{x}}=1}^{W}{\sum\limits_{{{r}_{y}}=1}^{H} \sum\limits_{{{r}_{z}}=1}^{D} \text{exp}( {V}_{j}^{\text{output}}(\vect{r})   )}} \label{eq:softmax2}\Big),
\end{equation}
and estimate the center of mass for each of the volumetric joint heatmaps to infer the positions of the joints in 3D: 
\begin{equation}
{\vect{y}_{j}}= \sum\limits_{{{r}_{x}}=1}^{W}{\sum\limits_{{{r}_{y}}=1}^{H} \sum\limits_{{{r}_{z}}=1}^{D} {\vect{r}\cdot {{V'}_{j}^{\text{output}}}(\vect{r})}} \label{eq:vol_center}
\end{equation}

Lifting to 3D allows getting more robust results, as the wrong predictions are spatially isolated from the correct ones inside the cube, so they can be discarded by convolutional operations. The network also naturally incorporates the camera parameters (uses them as an input), allows modelling the human pose prior and can reasonably handle multimodality in 2D detections.

\paragraph{Losses.} For both of the methods described above, the gradients pass from the output prediction of 3D joints' coordinates $\vect{y}_j$ to the input RGB-images $I_c$ making the pipeline trainable end-to-end. For the case of algebraic triangulation, we apply a soft version of per-joint Mean Square Error (MSE) loss to make the training more robust to outliers. This variant leads to better results compared to raw MSE or L1 (mean absolute error):
\begin{equation}
\mathcal{L}_j^{\text{alg}}(\theta,\phi)=
                \left\{
                \begin{array}{ll}
                  \text{MSE}(\vect{y}_j,\vect{y}^{\text{gt}}_j) \text{, if MSE}(\vect{y}_j,\vect{y}^{\text{gt}}_j)  < \varepsilon \\
                  \text{MSE}(\vect{y}_j,\vect{y}^{\text{gt}}_j)^{0.1} \cdot \varepsilon^{0.9} \text{, otherwise}\\
                \end{array}
              \right. \label{eq:loss_alg}
\end{equation}
Here, $\varepsilon$ denotes the threshold for the loss, which is set to $(20\text{ cm})^2$ in the experiments. The final loss is the average over all valid joints and all scenes in the batch.

For the case of volumetric triangulation, we use the L1 loss with a weak heatmap regularizer, which maximizes the prediction for the voxel that has inside of it the ground-truth joint:
\begin{equation}
\mathcal{L}^{\text{vol}}(\theta,\gamma,\nu)=
   \sum\limits_j{|\vect{y}_j-\vect{y}_j^{gt}| - \beta \cdot log(V_j^{\text{output}} (\vect{y}_j^{gt}))} \label{eq:loss_vol}
\end{equation}
 Without the second term, for some of the joints (especially, pelvis) the produced output volumetric heatmaps are not interpretable, probably due to insufficient size of the training datasets~\cite{sun2018integral}. Setting the $\beta$ to a small value ($\beta=0.01$) makes them interpretable, as the produced heatmaps always have prominent maxima close to the prediction. At the same time, such small $\beta$ does not seem to have any effect on  the final metrics, so its use can be avoided if interpretability is not needed. We have also tried the loss \eq{loss_alg} from the algebraic triangulation instead of L1, but it performed worse in our experiments.

\section{Experiments}

\paragraph{}
We conduct experiments on two available large multi-view datasets with available ground-truth 3D pose annotations: Human3.6M~\cite{h36m_pami} and CMU Panoptic~\cite{Joo_2017_TPAMI,xiang2018monocular,Simon_2017_CVPR}. 

\paragraph{Human3.6M dataset.} The Human3.6M~\cite{h36m_pami} is currently one of the largest 3D human pose benchmarks with many reported results both for monocular and multi-view setups. The full dataset consist of 3.6~million frames from 4 synchronized 50~Hz digital cameras along with the 3D pose annotations (collected using a marker-based MoCap system comprised of 10 separate IR-cameras). The dataset has 11 human subjects (5 females and 6 males) split into train, validation and test (only train and validation have the ground-truth annotations). 

The 2D backbone for Human3.6M was pretrained on the COCO dataset~\cite{lin2014microsoft} and finetuned jointly on MPII and Human3.6M for 10 epochs using the Adam optimizer with $10^{-4}$ learning rate. We use the 3D groundtruth and camera parameters provided by Martinez~\etal~\cite{martinez2017simple}. We undistort the images by applying grid-sampling on the video frames. If not mentioned explicitly, all networks are trained using four cameras and evaluated using all available cameras (either three or four, as one of the subjects lacks data from one camera). All algorithms use the 2D bounding boxes annotations provided with the dataset. The networks are trained for 6 epochs with $10^{-4}$ learning rate for the 2D backbone and a separate learning rate $10^{-3}$ for the volumetric backbone. 

Note that the volumetric triangulation approach uses predictions obtained from the algebraic triangulation (the whole system, however, potentially can be fine-tuned end-to-end). The size of volumetric cube $L$ was set to 2.5~m, which can enclose all subjects even if there is a few tens of centimeter error in pelvis prediction (which is much higher than the average error by the algebraic triangulation baseline). The number of output channels from the 2D backbone was set to $K=32$.  We did not apply any augmentations during the training, other than rotating the orientation of the cube in volumetric triangulation around the vertical axis.

\begin{table*}
  \resizebox{\textwidth}{!}{
   \begin{tabular}{lcccccccccccccccc}
    \toprule
    {Protocol 1 (relative to pelvis)}
    &  Dir. &  Disc. &  Eat &  Greet &  Phone &  Photo
    &  Pose &  Purch. &  Sit &  SitD. &  Smoke &  Wait
    &  WalkD. &  Walk &  WalkT. &  \textbf{Avg} \\
     \midrule
    Monocular methods (MPJPE relative to pelvis, mm)\\
     \midrule

    Martinez \etal \cite{martinez2017simple} &  51.8 &  56.2 &  58.1 &  59.0 &  69.5 &  78.4 &  55.2 &  58.1 &  74.0 &  94.6 &  62.3 &  59.1 &  65.1 &  49.5 &  52.4 &  62.9\\
    Sun~\etal\cite{sun2018integral} &  - &  - &  - &  - &  - &  - &  - &  - &  - &  - &  - &  - &  - &  - & - & \textbf{49.6}\\
    Pavllo~\etal~\cite{pavllo:videopose3d:2018} $(\ast)$ & 45.2 & 46.7 & 43.3 & 45.6 & 48.1 & 55.1 & 44.6 & 44.3 & 57.3 & 65.8 & 47.1 & 44.0 & 49.0 & 32.8 & 33.9 & \textbf{46.8} \\
    Hossain \& Little \cite{hossain2018exploiting} $(\ast)$ & 48.4 & 50.7 & 57.2 & 55.2 & 63.1 & 72.6 & 53.0 & 51.7 & 66.1 & 80.9 & 59.0 & 57.3 & 62.4 & 46.6 & 49.6 & 58.3 \\
    \textbf{Ours, volumetric single view ($\dagger$)} & 41.9 & 49.2 & 46.9 & 47.6 & 50.7 & 57.9 & 41.2 & 50.9 & 57.3 & 74.9 & 48.6 & 44.3 & 41.3 & 52.8 & 42.7 & 49.9\\
     \midrule
     \midrule
    Multi-view methods (MPJPE relative to pelvis, mm)\\
     \midrule
    Multi-View Martinez~\cite{Tome2018}  & 46.5 & 48.6 & 54.0 & 51.5 & 67.5 & 70.7 & 48.5 & 49.1 & 69.8 & 79.4 & 57.8 & 53.1 & 56.7 & 42.2 & 45.4 & 57.0 \\
    Pavlakos \etal \cite{Pavlakos2017} &  41.2 &  49.2 &  42.8 &  43.4 &  55.6 &  46.9 &  40.3 & 
    63.7 &  97.6 &  119.0 &  52.1 &  42.7 &  51.9 &  41.8 &  39.4 &  56.9\\
    Tome \etal~\cite{Tome2018} &  43.3 &  49.6 &  42.0 &  48.8 &  51.1 &  64.3 &  40.3 &  43.3 &  66.0 &  95.2 & 50.2 & 52.2 & 51.1 & 43.9 & 45.3 & 52.8\\
    Kadkhodamohammadi \& Padoy~\cite{Kadkhodamohammadi2018}& 39.4&46.9&41.0&42.7&53.6&54.8&41.4&50.0&59.9&78.8&49.8&46.2&51.1&40.5&41.0&49.1\\
    \hline
    RANSAC (our implementation) & 24.1 & 26.1 & 24.0 & 24.6 & 27.0 & 25.0 & 23.3 & 26.8 & 31.4 & 49.5 & 27.8 & 25.4 & 24.0 & 27.4 & 24.1 & 27.4\\    
    \textbf{Ours, algebraic (w/o conf)} & 22.9 & 25.3 & 23.7 & 23.0 & 29.2 & 25.1 & 21.0 & 26.2 & 34.1 & 41.9 & 29.2 & 23.3 & 22.3 & 26.6 & 23.3 & 26.9\\
    \textbf{Ours, algebraic} & 20.4 & 22.6 & 20.5 & 19.7 & 22.1 & 20.6 & 19.5 & 23.0 & 25.8 & 33.0 & 23.0 & 21.6 & 20.7 & 23.7 & 21.3 & 22.6\\
    \textbf{Ours, volumetric (softmax aggregation)} & \textbf{18.8} & \textbf{20.0} & 19.3 & 18.7 & \textbf{20.2} & \textbf{19.3} & 18.7 & 22.3 & 23.3 & 29.1 & \textbf{21.2} & \textbf{20.3} & \textbf{19.3} & \textbf{21.6} & \textbf{19.8} &\textbf{20.8}\\
    \textbf{Ours, volumetric (sum aggregation)} &  19.3 & 20.5 & 20.1 & 19.3 & 20.6 & 19.8 & 19.0 & 22.9 & 23.5 & 29.8 & 22.0 & 21.4 & 19.8 & 22.1 & 20.3 & 21.3\\
    \textbf{Ours, volumetric (conf aggregation)} &  19.9 & \textbf{20.0} & \textbf{18.9} & \textbf{18.5} & 20.5 & 19.4 & \textbf{18.4} & \textbf{22.1} & \textbf{22.5} &\textbf{ 28.7} & \textbf{21.2} & 20.8 & 19.7 & 22.1 & 20.2 & \textbf{20.8}\\
    
    \hline
    \bottomrule
   \end{tabular}
  }
 \caption {The results of evaluation on the Human3.6M dataset. The table presents the MPJPE error for the joints (relative to pelvis) for published state-of-the-art monocular and multi-view methods. The methods that are using temporal information during inference are marked by $(\ast)$. Note that our monocular method (labeled by $\dagger$) is using the approximate position of the pelvis estimated from the multi-view.
\label{table_hm36_1}}
\end{table*}

\begin{table*}
  \resizebox{\textwidth}{!}{
   \begin{tabular}{lcccccccccccccccc}
    \toprule
    {Protocol 1, absolute positions, filtered validation}
    &  Dir. &  Disc. &  Eat &  Greet &  Phone &  Photo
    &  Pose &  Purch. &  Sit &  SitD. &  Smoke &  Wait
    &  WalkD. &  Walk &  WalkT. &  {Avg} \\
     \midrule
    Multi-view methods (absolute MPJPE, mm)\\
     \midrule
    RANSAC & 21.6 & 22.9 & 20.9 & 21.0 & 23.1 & 23.0 & 20.8 & 22.0 & 26.4 & 26.6 & 24.0 & 21.5 & 21.0 & 23.9 & 20.8 & 22.8\\
    \textbf{Ours, algebraic (w/o conf)} & 21.7 & 23.7 & 22.2 & 20.4 & 26.7 & 24.2 & 19.9 & 22.6 & 31.2 & 35.6 & 26.8 & 21.2 & 20.9 & 24.6 & 21.1 & 24.5\\
    \textbf{Ours, algebraic} & 18.1 & 20.0 & 17.6 & 17.0 & 18.9 & 19.3 & 17.4 & 19.2 & 21.9 & 23.2 & 19.5 & 18.0 & 18.3 & 20.5 & 17.9 & 19.2\\
    \textbf{Ours, volumetric (softmax aggregation)} & \textbf{16.9} & \textbf{18.1} & 16.6 &\textbf{ 16.0} & \textbf{17.1} &\textbf{ 17.9} & \textbf{16.5} & \textbf{18.5} & 19.6 & \textbf{20.1} & \textbf{18.2} & \textbf{16.8} & \textbf{17.2} & \textbf{19.0} & \textbf{16.6} & \textbf{17.7}\\
    \textbf{Ours, volumetric (sum aggregation)}&  17.7 & 18.5 & 17.2 & 16.5 & 17.8 & 18.4 & 17.0 & 18.9 & 19.8 & 20.9 & 18.9 & 17.8 & 17.8 & 19.2 & 17.3 & 18.3  \\
    \textbf{Ours, volumetric (conf aggregation)} & 18.0 & 18.3 & \textbf{16.5} & 16.1 & 17.4 & 18.2 & \textbf{16.5} & \textbf{18.5} & \textbf{19.4} & \textbf{20.1} &\textbf{ 18.2} & 17.4 & \textbf{17.2} & 19.2 &\textbf{ 16.6} & 17.9\\
    
    \hline
    \bottomrule
   \end{tabular}
  }
 \caption {The results of evaluation on the Human3.6M dataset. The table presents the absolute positions MPJPE error for our algorithms. Note that the validation set has been filtered by removing the scenes with erroneous ground-truth 3D pose annotations.
\label{table_hm36_2}}
\end{table*}

For Human3.6M we follow the most popular protocol with 17-joint subset and testing on the validation. We used the MPJPE (Mean Per Joint Position Error) metric, which is L2 distance between the ground-truth and predicted positions of the joints (in most cases, measured with respect to pelvis). We use every fifth frame for the evaluation. As a baseline, we implemented a simple triangulation method with RANSAC and Huber loss, which is the de-facto standard for solving robust estimation problems. The baseline uses the same pretrained 2D backbone. The results of the standard protocol are summarized in Table \ref{table_hm36_1}.

Our implementations surpass the previous art by a large margin, even for the simplest RANSAC baseline. The introduced volumetric methods performs the best, providing about $30\%$ additional reduction in the error to the RANSAC, which is significant. 

While most of the works on monocular 3D human pose evaluate the positions of the joints relative to the pelvis (in order to avoid the estimation of global coordinates, which is problematic for monocular algorithms), evaluating with respect to the global coordinates is more reasonable for multi-view setups. This, however, is not straightforward due to 3D pose annotation errors in the Human3.6M - the problem is that some scenes of the 'S9' validation actor (parts of 'Greeting', 'SittingDown' and 'Waiting', see our project page) have the ground truth erroneously shifted in 3D compared to the actual position. Interestingly, the error is nullified when the pelvis is subtracted (as done for monocular methods), however, to make the results for the multi-view setup interpretable we must exclude these scenes from the evaluation. The results for the absolute MPJPE for our methods with these scenes excluded are presented in Table \ref{table_hm36_2}, giving a better sense of the magnitude of errors. Interestingly, the average MPJPE for volumetric is much smaller than the size of the voxel (3.9 cm), showing the importance of the subpixel resolving soft-argmax.

Our volumetric multi-view methods can be generalized to the case of a single camera view, naturally taking into account the camera intrinsics. To check the monocular performance we have done a separate experiment with training using a random number of cameras from 1 to 4. For the case of a single camera we used the L1 loss on joint positions relative to the pelvis (as is usually done for the monocular methods). Without any tuning this resulted in 49.9~mm error, which is close to the current state of the art. Note that the position of the cube center was estimated by triangulating the pelvis from all 4 cameras, so the performance of the methods might be somewhat overoptimistic. On the other hand, the average relative positions error of the monocular method lies within 4-6 cm, which corresponds to a negligible shift for the volumetric cube position (less than 2 voxels).

\begin{figure*}

    \centering
    \includegraphics[width=0.95\textwidth]{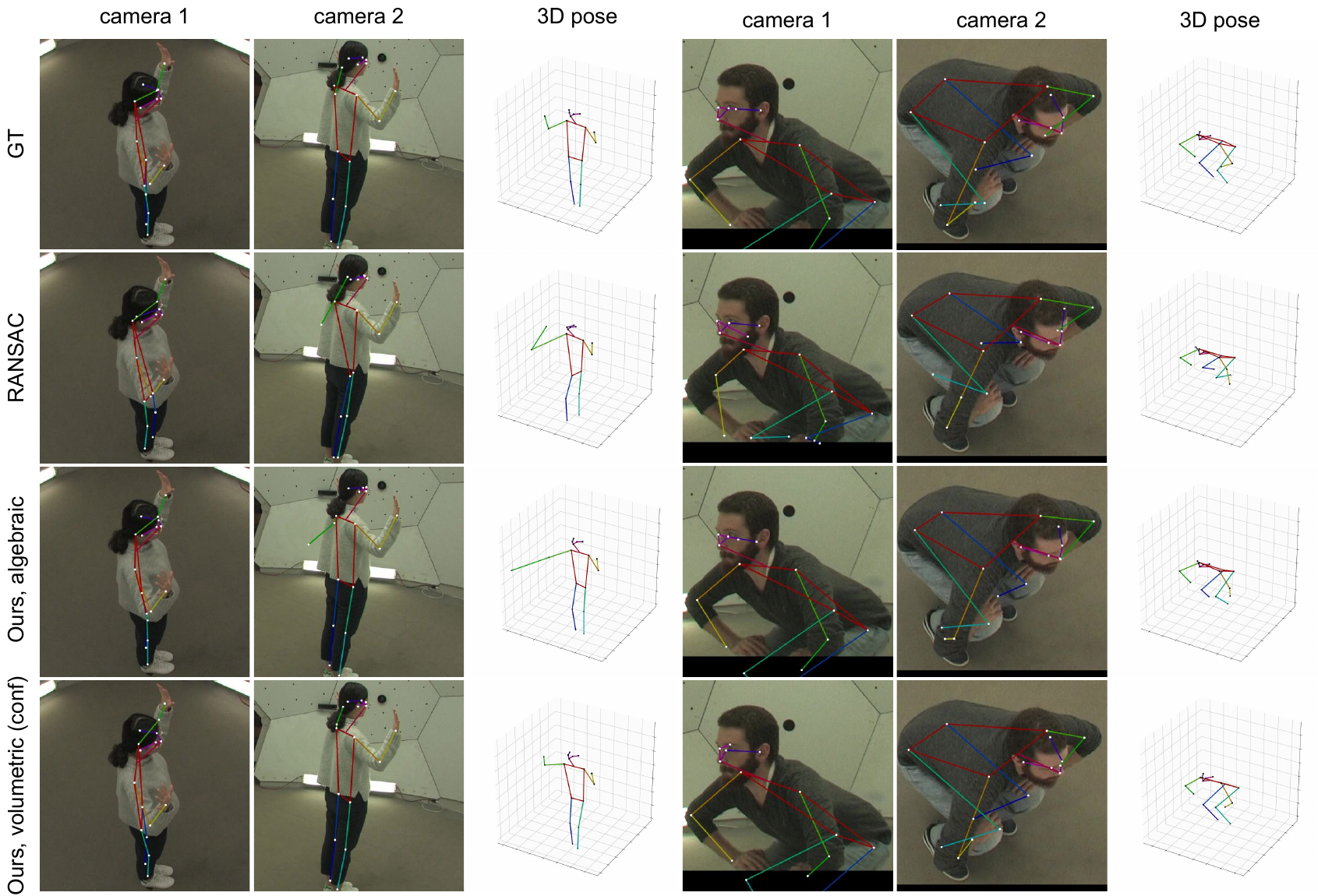}
    \caption{Illustration of the difference in performance of the approaches on the CMU dataset validation (using 2 cameras) that demonstrates the robustness of the volumetric triangulation approach.}
    \label{fig:cmu_show}
    
\end{figure*}

\begin{figure}

    \centering
    \includegraphics[width=0.45\textwidth]{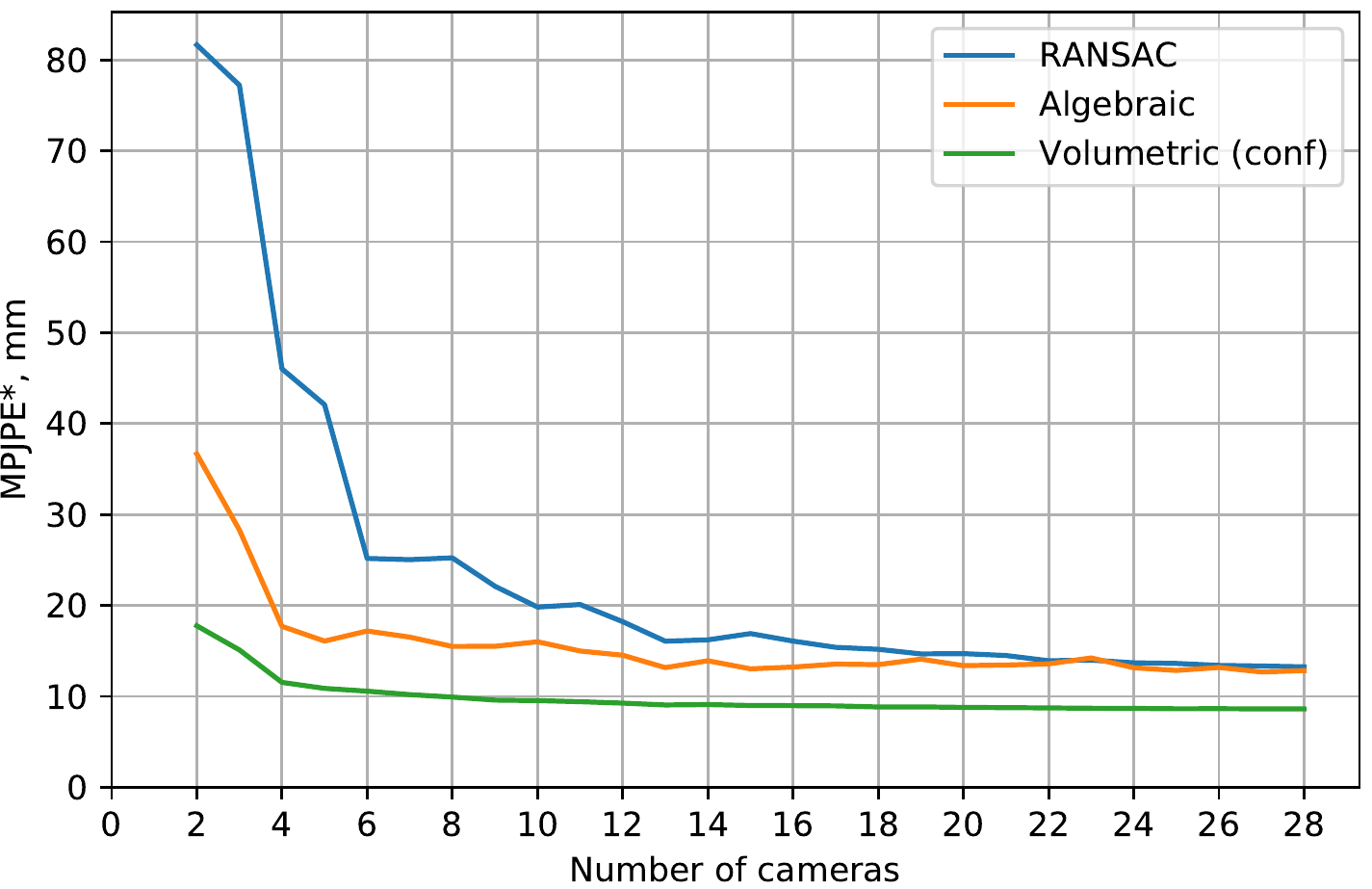}
    \caption{Estimate of the MPJPE absolute error on the subset of CMU validation versus the numbers of cameras (up to 28, treating the annotations from CMU as ground truth). Each value on the plot is obtained by sampling 50 random subsets of cameras followed by averaging.}
    \label{fig:cam_scaling}
    
\end{figure}

\begin{table}
	\centering
	\small
	\resizebox{0.5\textwidth}{!}{%
		\begin{tabular}{{p{8cm}p{2cm}}}
		\toprule
		Model &  MPJPE, mm \\
		\midrule
		RANSAC & 39.5\\
        \textbf{Ours, algebraic (w/o conf)} & 33.4\\
        \textbf{Ours, algebraic} & 21.3\\
        \textbf{Ours, volumetric (softmax aggregation)} & \textbf{13.7}\\
        \textbf{Ours, volumetric (sum aggregation)} & \textbf{13.7}\\
        \textbf{Ours, volumetric (conf aggregation)} & 14.0\\
		\bottomrule
		\end{tabular}
	}
	\caption{Results of evaluation on the CMU dataset in terms of MPJPE error on the CMU dataset validation (using 4 cameras).}
	\label{table_cmu_1}
\end{table}

\begin{figure*}

    \centering
    \includegraphics[width=0.8\textwidth]{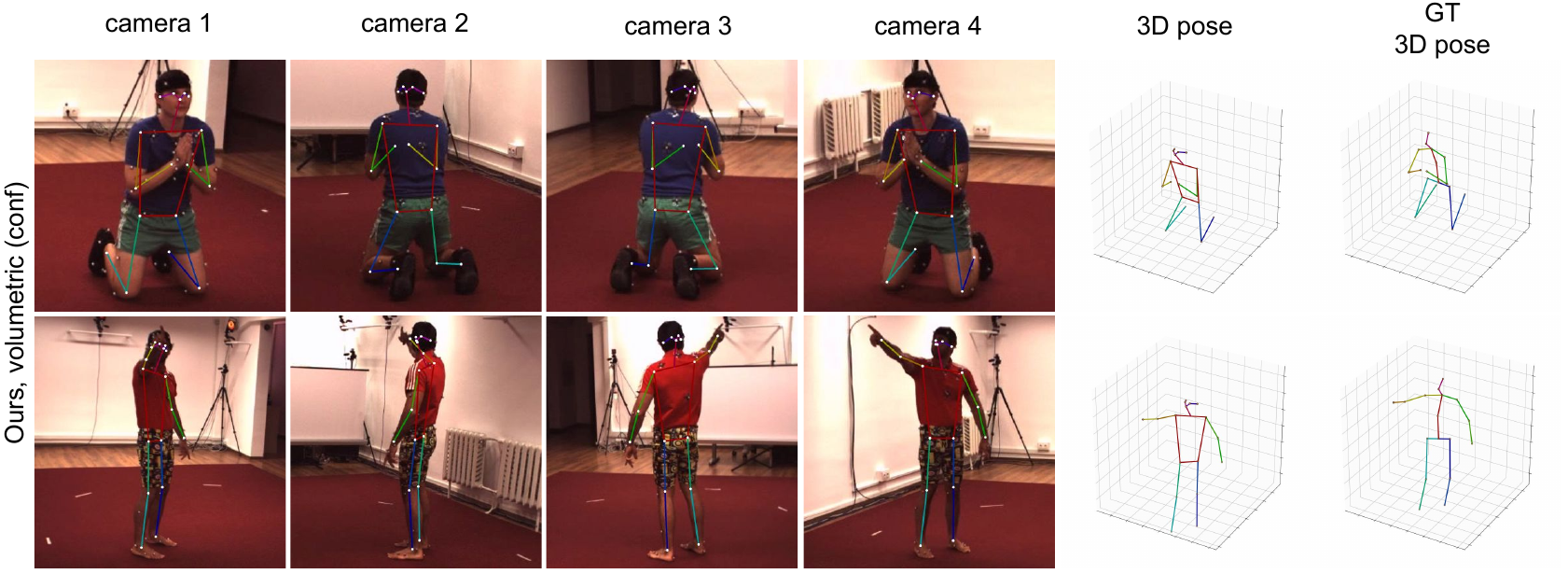}
    \caption{Demonstration of successful transfer of the solution trained on CMU dataset to Human3.6M scenes. Note that keypoint skeleton models on Human3.6M and CMU are different.}
    \label{fig:cmu_transfer}
    
\end{figure*}

\paragraph{CMU Panoptic dataset.} The CMU panoptic is a new multi-camera dataset maintained by the Carnegie Mellon University~\cite{Joo_2017_TPAMI,xiang2018monocular,Simon_2017_CVPR}. The dataset provides 30~Hz Full-HD videostreams of 40 subjects from up to 31 synchronized cameras.

The dataset is provided with annotations of the Full-HD cameras acquired via triangulation using all camera views. Since there are no other published results on the quality of multi-view pose estimation on CMU, we use our own protocol. For the tests we use the 17-subset of the 19-joint annotation format in the dataset, which coincides with the popular COCO format~\cite{lin2014microsoft}. We used the same train/val split as in \cite{xiang2018monocular}, which only has scenes with a single person at each point of time. Additionally, we split the dataset by camera views (4 cameras in val, up to 27 cameras in train), so there is no overlap between the test and validation both in terms of subjects and camera views. To get the human bounding boxes we use Mask R-CNN 2D detector with ResNet-152 backbone~\cite{massa2018mrcnn}). Note that we process the frames without taking into account the lens distortion which leads to some loss of accuracy. 

The networks are trained using the Adam optimizer similarly to Human3.6M as described in the previous section. The number of views during training was selected randomly from 2 to 5.

The comparison between our multi-view methods is presented in Table \ref{table_cmu_1} with absolute MPJPE used as the main metric. Here, the volumetric approach has a dramatic advantage over the algebraic one, and its superiority is far more evident than on Human3.6M. We believe that the main point in which CMU differs from Human3.6M is that in CMU most cameras do not have the full view of a person the whole time, leading to strong occlusions and missing parts. This suggests the importance of the 3D prior that can be learnt only by volumetric models. It seems that RANSAC performs worse compared to algebraic triangulation without confidences because it is not finetuned on the CMU data. The difference between the methods' prediction is illustrated in \fig{cmu_show}. We have not observed any significant difference between the volumetric aggregation methods.

In \fig{cam_scaling} we present a plot for the error versus the numbers of used cameras. The plot demonstrates that the proposed volumetric triangulation methods allow drastically reducing the number of cameras in real-life setups: the accuracy on "meta"-groundtruth for RANSAC approach with 28 cameras is surpassed by the volumetric approach with just four cameras.

We also conducted experiments to demonstrate that the learnt model indeed generalizes to new setups. For that we applied a CMU-trained model to Human3.6M validation scenes. The qualitative results for the learnable triangulation are presented in \fig{cmu_transfer}. Please see videos for all of the methods on our project page. To provide a quantitative measure of the generalizing ability, we have measured the MPJPE for the set of joints which seem to have the most similar semantics (namely, 'elbows', 'wrists' and 'knees'). The measured MPJPE is 36 mm for learnable triangulation and 34 mm for the volumetric approach, which seems reasonable when compared to the results of the methods trained on Human3.6M (16-18 mm, depending on the triangulation method). 
\section{Conclusion}
We have presented two novel methods for the multi-view 3D human pose estimation based on learnable triangulation that achieve state-of-the-art performance on the Human3.6M dataset. The proposed solutions drastically reduce the number of views needed to achieve high accuracy, and produce smooth pose sequences on the CMU Panoptic dataset without any temporal processing (see our project page for demonstration), pointing that it can potentially improve the ground truth annotation of the dataset. An ability to transfer the trained method between setups is demonstrated for the CMU Panoptic $\rightarrow$ Human3.6M pair.  
The volumetric triangulation strongly outperformed all other approaches both on CMU Panoptic and Human3.6M datasets. We speculate that due to its ability to learn a human pose prior this method is robust to occlusions and partial views of a person. Another important advantage of this method is that it explicitly takes the camera parameters as independent input. Finally, volumetric triangulation also generalizes to \textit{monocular} images if human's approximate position is known, producing results close to state of the art.

One of the major limitations of our approach is that it supports only a single person in the scene. This problem can be mitigated by applying available ReID solutions to the 2D detections of humans, however there might be more seamless solutions. Another major limitation of the volumetric triangulation approach is that it relies on the predictions of the algebraic triangulation. This leads to the need for having at least two camera views that observe the pelvis, which might be a problem for some applications. The performance of our method can also potentially be further improved by adding multi-stage refinement in a way similar to \cite{Tome2018}.

\FloatBarrier
\ifnum\value{page}>8 \errmessage{Number of pages exceeded!!!!}\fi

{\small
\bibliographystyle{ieee}
\bibliography{refs}

\begin{thebibliography}{10}\itemsep=-1pt

\bibitem{hartley2003multiple}
R.~Hartley and A.~Zisserman.
\newblock {\em Multiple view geometry in computer vision}.
\newblock Cambridge university press, 2003.

\bibitem{hossain2018exploiting}
M.~R.~I. Hossain and J.~J. Little.
\newblock Exploiting temporal information for 3d human pose estimation.
\newblock In {\em European Conference on Computer Vision}, pages 69--86.
  Springer, 2018.

\bibitem{h36m_pami}
C.~Ionescu, D.~Papava, V.~Olaru, and C.~Sminchisescu.
\newblock Human3.6m: Large scale datasets and predictive methods for 3d human
  sensing in natural environments.
\newblock {\em IEEE Transactions on Pattern Analysis and Machine Intelligence},
  36(7):1325--1339, jul 2014.

\bibitem{jaderberg2015spatial}
M.~Jaderberg, K.~Simonyan, A.~Zisserman, et~al.
\newblock Spatial transformer networks.
\newblock In {\em Advances in neural information processing systems}, pages
  2017--2025, 2015.

\bibitem{Joo_2017_TPAMI}
H.~Joo, T.~Simon, X.~Li, H.~Liu, L.~Tan, L.~Gui, S.~Banerjee, T.~S. Godisart,
  B.~Nabbe, I.~Matthews, T.~Kanade, S.~Nobuhara, and Y.~Sheikh.
\newblock Panoptic studio: A massively multiview system for social interaction
  capture.
\newblock {\em IEEE Transactions on Pattern Analysis and Machine Intelligence},
  2017.

\bibitem{Kadkhodamohammadi2018}
A.~Kadkhodamohammadi and N.~Padoy.
\newblock {A generalizable approach for multi-view 3D human pose regression}.
\newblock apr 2018.

\bibitem{lin2014microsoft}
T.-Y. Lin, M.~Maire, S.~Belongie, J.~Hays, P.~Perona, D.~Ramanan,
  P.~Doll{\'a}r, and C.~L. Zitnick.
\newblock Microsoft coco: Common objects in context.
\newblock In {\em European conference on computer vision}, pages 740--755.
  Springer, 2014.

\bibitem{martinez2017simple}
J.~Martinez, R.~Hossain, J.~Romero, and J.~J. Little.
\newblock A simple yet effective baseline for 3d human pose estimation.
\newblock In {\em Proceedings of the IEEE International Conference on Computer
  Vision}, pages 2640--2649, 2017.

\bibitem{massa2018mrcnn}
F.~Massa and R.~Girshick.
\newblock {maskrcnn-benchmark: Fast, modular reference implementation of
  Instance Segmentation and Object Detection algorithms in PyTorch}.
\newblock \url{https://github.com/facebookresearch/maskrcnn-benchmark}, 2018.
\newblock Accessed: 01 Feb 2019.

\bibitem{moeslund2006survey}
T.~B. Moeslund, A.~Hilton, and V.~Kr{\"u}ger.
\newblock A survey of advances in vision-based human motion capture and
  analysis.
\newblock {\em Computer vision and image understanding}, 104(2-3):90--126,
  2006.

\bibitem{Moon_2018_CVPR_V2V-PoseNet}
G.~Moon, J.~Chang, and K.~M. Lee.
\newblock V2v-posenet: Voxel-to-voxel prediction network for accurate 3d hand
  and human pose estimation from a single depth map.
\newblock In {\em The IEEE Conference on Computer Vision and Pattern
  Recognition (CVPR)}, 2018.

\bibitem{Pavlakos2017}
G.~Pavlakos, X.~Zhou, K.~G. Derpanis, and K.~Daniilidis.
\newblock {Harvesting multiple views for marker-less 3D human pose
  annotations}.
\newblock {\em Proceedings - 30th IEEE Conference on Computer Vision and
  Pattern Recognition, CVPR 2017}, 2017-Janua:1253--1262, 2017.

\bibitem{pavllo:videopose3d:2018}
D.~Pavllo, C.~Feichtenhofer, D.~Grangier, and M.~Auli.
\newblock 3d human pose estimation in video with temporal convolutions and
  semi-supervised training.
\newblock {\em arXiv}, abs/1811.11742, 2018.

\bibitem{Rhodin2018}
H.~Rhodin, F.~Meyer, J.~Sporri, E.~Muller, V.~Constantin, P.~Fua,
  I.~Katircioglu, and M.~Salzmann.
\newblock {Learning Monocular 3D Human Pose Estimation from Multi-view Images}.
\newblock In {\em 2018 IEEE/CVF Conference on Computer Vision and Pattern
  Recognition}, pages 8437--8446. IEEE, jun 2018.

\bibitem{Simon_2017_CVPR}
T.~Simon, H.~Joo, and Y.~Sheikh.
\newblock Hand keypoint detection in single images using multiview
  bootstrapping.
\newblock {\em CVPR}, 2017.

\bibitem{sun2018integral}
X.~Sun, B.~Xiao, F.~Wei, S.~Liang, and Y.~Wei.
\newblock Integral human pose regression.
\newblock In {\em Proceedings of the European Conference on Computer Vision
  (ECCV)}, pages 529--545, 2018.

\bibitem{tome2017lifting}
D.~Tome, C.~Russell, and L.~Agapito.
\newblock Lifting from the deep: Convolutional 3d pose estimation from a single
  image.
\newblock In {\em Proceedings of the IEEE Conference on Computer Vision and
  Pattern Recognition}, pages 2500--2509, 2017.

\bibitem{Tome2018}
D.~Tome, M.~Toso, L.~Agapito, and C.~Russell.
\newblock {Rethinking Pose in 3D: Multi-stage Refinement and Recovery for
  Markerless Motion Capture}.
\newblock In {\em 2018 International Conference on 3D Vision (3DV)}, pages
  474--483. IEEE, sep 2018.

\bibitem{von2016human}
T.~Von~Marcard, G.~Pons-Moll, and B.~Rosenhahn.
\newblock Human pose estimation from video and imus.
\newblock {\em IEEE transactions on pattern analysis and machine intelligence},
  38(8):1533--1547, 2016.

\bibitem{xiang2018monocular}
D.~Xiang, H.~Joo, and Y.~Sheikh.
\newblock Monocular total capture: Posing face, body, and hands in the wild.
\newblock {\em arXiv preprint arXiv:1812.01598}, 2018.

\bibitem{xiao2018simple}
B.~Xiao, H.~Wu, and Y.~Wei.
\newblock Simple baselines for human pose estimation and tracking.
\newblock In {\em European Conference on Computer Vision (ECCV)}, 2018.

\bibitem{yu2018humbi}
Z.~Yu, J.~S. Yoon, P.~Venkatesh, J.~Park, J.~Yu, and H.~S. Park.
\newblock Humbi 1.0: Human multiview behavioral imaging dataset.
\newblock {\em arXiv preprint arXiv:1812.00281}, 2018.

\end{thebibliography}
}
\end{document}